\documentclass[letterpaper]{article} 
\usepackage[draft]{aaai2026}
\usepackage{times}  
\usepackage{helvet}  
\usepackage{courier}  
\usepackage[hyphens]{url}  
\usepackage{graphicx} 
\usepackage{natbib}  
\usepackage{caption} 
\usepackage{algorithm}
\usepackage{algorithmic}
\usepackage{booktabs}
\usepackage[table,xcdraw]{xcolor}  
\usepackage{amssymb}  
\usepackage{amsmath}
\usepackage{multirow}

\usepackage{newfloat}
\usepackage{listings}
\DeclareCaptionStyle{ruled}{labelfont=normalfont,labelsep=colon,strut=off} 
\floatstyle{ruled}
\newfloat{listing}{tb}{lst}{}
\floatname{listing}{Listing}
\title{PointLAMA: Latent Attention meets Mamba for Efficient Point Cloud Pretraining}

\newcommand{\ourmethod}{PointLAMA}

\author{
    Xuanyu Lin\textsuperscript{\rm 1} 
    Xiaona Zeng\textsuperscript{\rm 2}
    Xianwei Zheng\textsuperscript{\rm 2}
    \thanks{Corresponding Author.}
    Xutao Li\textsuperscript{\rm 1}\\
}
\affiliations{
    \textsuperscript{\rm 1} Shantou University
    \textsuperscript{\rm 2} Foshan University \\
23xylin@stu.edu.cn, alex.w.zheng@hotmail.com \\
}

\usepackage{bibentry}

\begin{document}

\maketitle

\begin{abstract}
Mamba has recently gained widespread attention as a backbone model for point cloud modeling, leveraging a state-space architecture that enables efficient global sequence modeling with linear complexity. However, its lack of local inductive bias limits its capacity to capture fine-grained geometric structures in 3D data. To address this limitation, we propose \textbf{PointLAMA}, a point cloud pretraining framework that combines task-aware point cloud serialization, a hybrid encoder with integrated Latent Attention and Mamba blocks, and a conditional diffusion mechanism built upon the Mamba backbone. Specifically, the task-aware point cloud serialization employs Hilbert/Trans-Hilbert space-filling curves and axis-wise sorting to structurally align point tokens for classification and segmentation tasks, respectively. Our lightweight Latent Attention block features a Point-wise Multi-head Latent Attention (PMLA) module, which is specifically designed to align with the Mamba architecture by leveraging the shared latent space characteristics of PMLA and Mamba. This enables enhanced local context modeling while preserving overall efficiency. To further enhance representation learning, we incorporate a conditional diffusion mechanism during pretraining, which denoises perturbed feature sequences without relying on explicit point-wise reconstruction. Experimental results demonstrate that PointLAMA achieves competitive performance on multiple benchmark datasets with minimal parameter count and FLOPs, validating its effectiveness for efficient point cloud pretraining.
\end{abstract}


\section{Introduction}

Point cloud understanding is fundamental in various 3D vision applications, such as autonomous driving, robotics, and augmented reality. With the rise of transformer-based models like Point Transformer~\cite{zhao2021point}, remarkable progress has been made in point cloud classification and segmentation tasks. However, the quadratic complexity of the attention mechanism in transformers limits their scalability and efficiency, particularly in high-resolution and real-time scenarios.

To overcome this issue, state space models (SSMs) have emerged as a promising alternative. Mamba, a hardware-efficient SSM, enables linear-time global modeling by replacing attention with selective recurrence. PointMamba~\cite{liang2024pointmamba} adapts this to point cloud analysis by transforming unordered point clouds into serialized sequences using space-filling curves, and applying a plain Mamba encoder. Despite its impressive efficiency, PointMamba~\cite{liang2024pointmamba} struggles with modeling local geometric structures due to the lack of local inductive bias inherent in unidirectional SSMs.

Meanwhile, generative pretraining has become a powerful paradigm in 3D learning. Traditional reconstruction-based methods like Point-BERT~\cite{yu2022point} and Point-MAE~\cite{pang2022masked} rely on computationally expensive losses like Chamfer Distance~\cite{fan2017point} or Earth Mover's Distance~\cite{rubner2000earth}, which often produce blurry outputs due to the unordered nature of point clouds. Recently, diffusion-based pretraining methods, such as PointDif~\cite{zheng2024pointdif}, have demonstrated superior performance by learning to denoise corrupted point clouds via conditional generative processes, effectively capturing both global context and local structures. However, their reliance on Transformer architectures leads to high computational complexity.

\begin{figure*}[htbp]
	\begin{center}
		\includegraphics[width=0.8\linewidth]{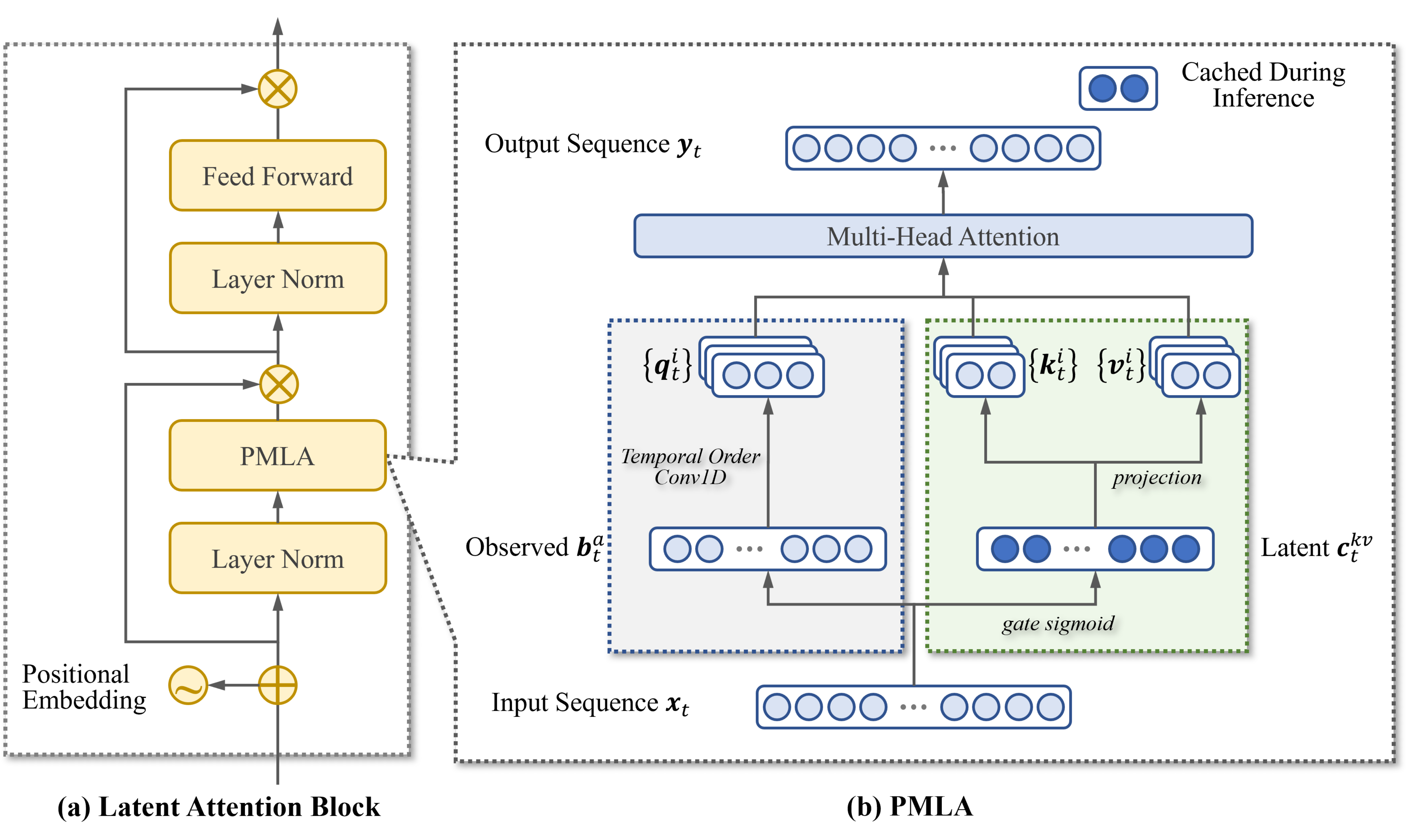}
	\end{center}
	\caption{Schematic illustration of our Latent Attention Block. (a) A latent attention block consists of PMLA, layer normalization, residual connections, and a feed-forward layer. (b) Internal design of our PMLA module with temporal Conv1D and gated latent projection.}
	\label{fig:mla_block}
\end{figure*}

To address the limitations of existing methods and unify the advantages of state space modeling and generative pretraining, we propose \textbf{PointLAMA}, a \emph{Point-wise information fusion with Multi-head Latent Attention (PMLA) and Mamba-based architecture}. Our framework introduces a lightweight point-wise latent attention module called \emph{PMLA} into the Mamba pipeline, enhancing local context modeling with minimal computational overhead. Furthermore, we design task-specific point serialization strategies to convert unordered point sets into structure-aware sequences suitable for SSMs: space-filling curves (Hilbert and Trans-Hilbert) for classification, and axis-wise sorting for segmentation. To further improve representation learning, we incorporate a conditional diffusion-based pretraining scheme that perturbs encoded features and progressively restores geometric structure via a denoising network, avoiding explicit point set reconstruction.

In conclusion, the contributions of this paper are threefold: 
1) We propose \textbf{PointLAMA}, a hybrid architecture integrating PMLA with a Mamba-based state space model, enabling both efficient global modeling and lightweight local feature enhancement in point cloud sequences.
2) We design task-specific point serialization strategies, including Hilbert-based and axis-wise sorting methods, to transform unordered 3D points into structure-aware sequences suitable for state space modeling.
3) We develop a feature-space conditional diffusion pretraining framework, which denoises perturbed point features without explicit geometric reconstruction, thus enhancing geometric representation learning with lower computational cost.

\section{Related Work}

\noindent \textbf{Fusion of Self-Attention and State Space Models (SSM).} Self-attention~\cite{vaswani2017attention} mechanisms have been widely adopted in point cloud analysis for their ability to capture long-range dependencies across unordered 3D points. A representative method, Point Transformer~\cite{guo2021pct, zhao2021point}, employs attention with positional encoding to aggregate local geometric features and achieve strong performance on various 3D tasks. However, its quadratic computational complexity with respect to the number of points ($O(N^2)$) poses challenges for scalability and efficiency in large-scale or real-time applications.

To address these challenges, state space models (SSMs)~\cite{gu2022parameterization, gupta2022diagonal, gu2022train} have emerged as efficient alternatives with linear time complexity ($O(N)$). Representative approaches include Point Cloud Mamba (PCM)~\cite{zhang2024point}, which adapts Mamba-based SSMs to point cloud classification~\cite{uy2019revisiting}; Serialized Point Mamba, which leverages space-filling curves to transform unordered points into sequential inputs compatible with SSMs; and PointMamba~\cite{liang2024pointmamba}, which improves structural scalability and hierarchical representation through compact state propagation. These methods validate the potential of SSMs for efficient global modeling. However, despite their theoretical advantages, Mamba-based architectures are currently less hardware-friendly than CNNs and Transformers, as they do not fully exploit matrix multiplication accelerators (e.g., GPUs, TPUs), resulting in practical performance inefficiencies.

\begin{figure*}
	\begin{center}
		\includegraphics[width=1.0\linewidth]{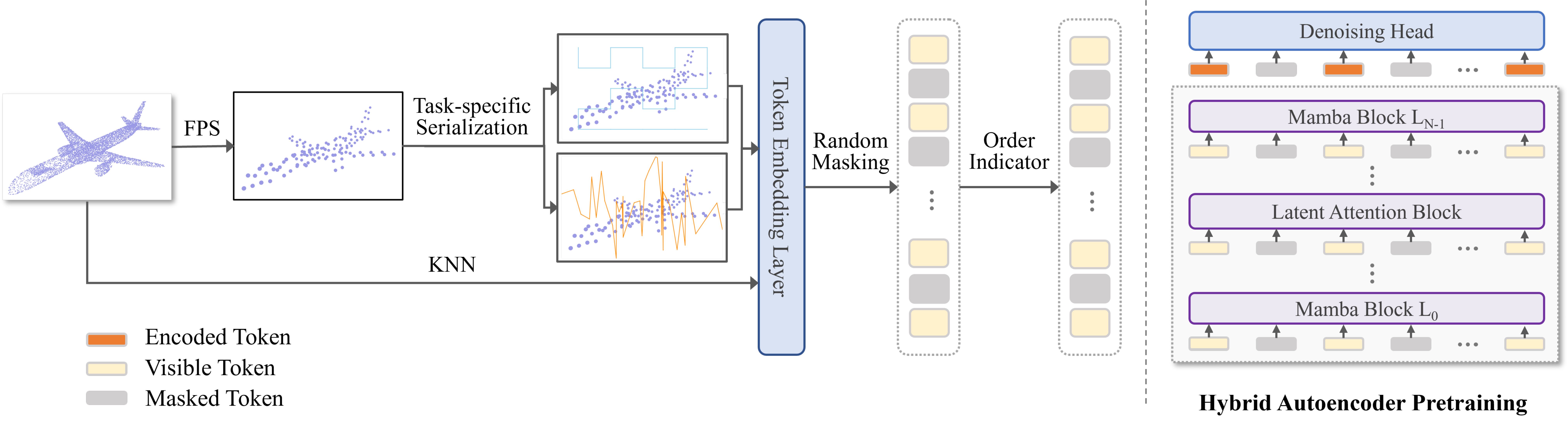}
	\end{center}
	\caption{Pretraining pipeline of our PointLAMA. The framework consists of two stages, separated by a dashed line. The upper stage processes the input point cloud via FPS-based grouping, task-specific serialization (e.g., Hilbert or axis-based), masking, and order embedding. The lower stage feeds the masked tokens into a hybrid autoencoder architecture, which comprises stacked Mamba Blocks, a Latent Attention Block (see Fig.~\ref{fig:mla_block}), and a denoising head derived from a conditional diffusion model, used for noise prediction and representation learning.}
    \label{fig:pipeline}
\end{figure*}

\noindent \textbf{Diffusion-Based Pretraining for Point Clouds.} Early point cloud pretraining~\cite{poursaeed2020self, qi2023contrast} methods often relied on explicit reconstruction, where a subset of points was randomly masked and the model was trained to predict the missing geometry. Representative frameworks such as Point-BERT~\cite{yu2022point}, Point-MAE~\cite{pang2022masked}, and Point-M2AE~\cite{zhang2022point} adopted masked completion strategies to encourage the model to capture both local and global geometric structures. These methods are typically supervised using reconstruction losses such as Chamfer Distance (CD)~\cite{fan2017point} or Earth Mover’s Distance (EMD)~\cite{rubner2000earth}, which measure the geometric discrepancy between predicted and ground-truth point sets. However, such losses require generating full point sets and involve costly operations like bidirectional nearest-neighbor or optimal transport computations. This not only increases computational overhead but also leads to blurry or inconsistent reconstructions due to the unordered and structureless nature of point clouds.

Diffusion models~\cite{sohl2015deep} provide an alternative generative pretraining paradigm that avoids explicit point set reconstruction. Instead, they learn to recover the original data distribution through a gradual denoising process from noisy or perturbed inputs. PointDif~\cite{zheng2024pointdif} is one of the earliest attempts to incorporate conditional diffusion into point cloud pretraining, where the denoising process is guided by high-level features extracted from an encoder. PVD~\cite{zhou2021point, zhou2023pvd} further refined this idea by modeling diffusion at the individual point level, enhancing the model’s sensitivity to local geometric patterns. Compared to traditional methods, diffusion-based approaches better accommodate the irregular and unordered structure of point clouds, while demonstrating stronger generalization and stability in downstream tasks.

\section{Preliminaries}
\label{sec:Preliminaries}
\textbf{State Space Model and Multi-head Latent Attention.} A state space model (SSM) describes the evolution of a latent state $h_t \in \mathbb{R}^N$ over time based on input $x_t \in \mathbb{R}^D$:
\begin{equation}
    h_t = A h_{t-1} + B x_t,\quad y_t = C h_t + D x_t,
\end{equation}
where $A, B, C, D$ are learnable parameters. The Mamba~\cite{gu2023mamba} architecture extends this formulation by introducing input-dependent $B$, $C$, and $\Delta$, resulting in a selective SSM that adapts its dynamics at each timestep. To ensure efficiency, Mamba uses a hardware-aware recurrent scan algorithm, achieving linear time complexity in sequence length.

Meanwhile, Multi-head Latent Attention (MLA) is a lightweight attention mechanism that reduces memory and computation by replacing full-rank key and value projections with low-rank decompositions. Given input $X \in \mathbb{R}^{T \times D}$, MLA computes:
\begin{equation}
    Q = X W_Q,\quad K = X W_K^a W_K^b,\quad V = X W_V^a W_V^b,
\end{equation}
where $W_Q \in \mathbb{R}^{D \times n_h d_h}$, $W_K^a, W_V^a \in \mathbb{R}^{D \times r}$, and $W_K^b, W_V^b \in \mathbb{R}^{r \times n_h d_h}$. After splitting $Q, K, V$ into $n_h$ heads, attention is computed as:
\begin{equation}
    O = \sum_{i=1}^{n_h} \mathrm{softmax} \left( \frac{Q_i K_i^\top}{\sqrt{d_h}} \right) V_i W_{O,i},
\end{equation}
where $W_{O,i} \in \mathbb{R}^{d_h \times D}$. MLA significantly reduces KV cache size while maintaining expressiveness, and integrates well with state space architectures like Mamba.

\noindent \textbf{Point Token Serialization}. A point cloud can be represented as a set of local feature vectors in low-dimensional space, referred to as \textit{point tokens}. Unlike tokens in images or text, point clouds are inherently unordered and lack a natural sequential structure, making them incompatible with sequence-based models such as Mamba.

To bridge this gap, we propose task-specific serialization strategies. For classification tasks, we adopt Hilbert and Trans-Hilbert space-filling curves to construct globally continuous sequences that preserve spatial coherence. For segmentation tasks, we design a coordinate-wise sorting and concatenation strategy to emphasize local neighborhood information. These serialization schemes inject structural priors into the model, enabling Mamba to effectively capture spatial dependencies in point clouds.

\section{Method}
We propose \textbf{PointLAMA}, a hybrid pretraining framework for point clouds, as illustrated in Figure~\ref{fig:pipeline}. The overall architecture integrates Point-wise Multi-head Latent Attention (PMLA) with the Mamba state space model. To handle the unordered and irregular nature of point clouds, PointLAMA first transforms raw 3D data into structured 1D sequences via a task-aware pipeline. Specifically, it applies Farthest Point Sampling (FPS) and k-Nearest Neighbors (KNN) to form local patches, encodes each patch with a lightweight PointNet, and then serializes the resulting tokens using either space-filling curves or axis-wise sorting depending on downstream tasks. The serialized sequence is partially masked and injected with order embeddings before being fed into our hybrid encoder, which consists of stacked Mamba blocks and a Latent Attention Block. During pretraining, a conditional diffusion model perturbs the visible features and guides the decoder to restore the masked ones, enabling the encoder to capture structural representations without explicit point-wise supervision.

\subsection{Task-Aware Point Cloud Serialization}
To adapt point clouds to the 1D sequential input format required by Mamba, we propose a task-aware serialization pipeline. It first encodes local patch features and then arranges them into ordered sequences using strategies tailored to classification or segmentation tasks. To simulate partial observation during pretraining, a subset of tokens is masked and excluded from the encoder input. Additionally, learnable order embeddings are introduced to differentiate serialization patterns and enhance spatial adaptability.

\noindent \textbf{Grouping and Local Feature Extraction.}
We first apply Farthest Point Sampling (FPS) and k-Nearest Neighbors (KNN) to partition the input point cloud into $G$ local patches.
Each patch is then encoded by a lightweight PointNet~\cite{qi2017pointnet++} to produce center-aligned feature tokens $F \in \mathbb{R}^{B \times G \times C}$.

\noindent \textbf{Task-Specific Serialization and Alignment.} To transform unordered token sets into sequences suitable for Mamba, we apply two complementary serialization strategies depending on the target task. Importantly, the sorting is first applied to the center points $C$, and the same order is used to realign their associated neighborhood patches $N$, ensuring consistency between spatial locality and token order.

For shape classification tasks, we aim to preserve global structure. We adopt Hilbert or Trans-Hilbert space-filling curves~\cite{moon2001analysis} to generate a 1D ordering over $C$. Specifically, we normalize the 3D coordinates of center points into a discrete grid, compute their Hilbert indices, and then sort them accordingly. The resulting order is used to rearrange the center tokens and flatten the neighborhood tensor as:
\begin{equation}
    N' = \text{reshape}(N_{\texttt{order}}) \in \mathbb{R}^{B \times G \times S \times 3}.
\end{equation}
This ordering ensures that nearby points in space are also close in sequence, improving global modeling capability under the Mamba framework.

\begin{figure}
	\begin{center}
		\includegraphics[width=1.0\linewidth]{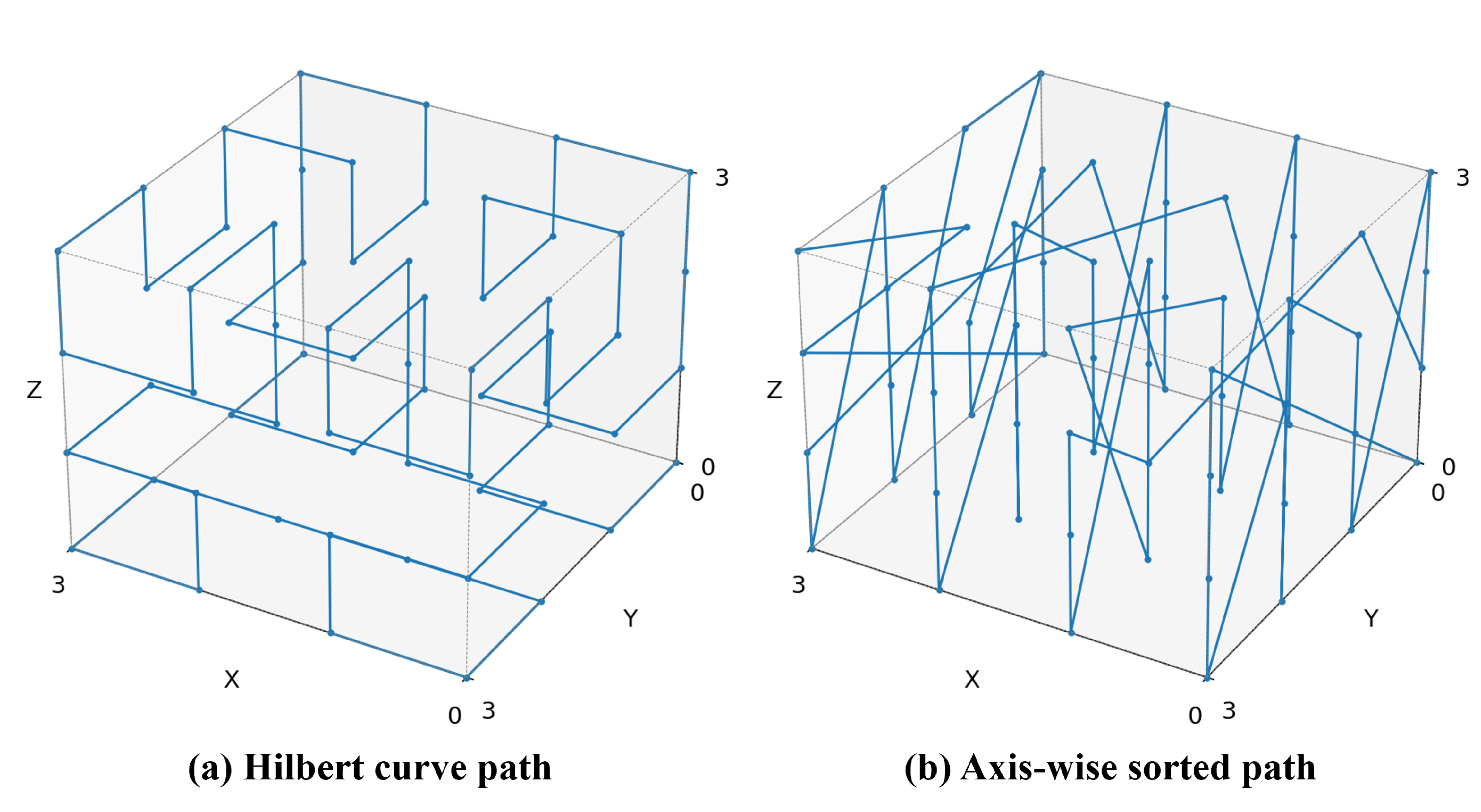}
	\end{center}
	\caption{Visualization of our two serialization strategies for point clouds. Left: Hilbert space-filling curve ordering used in classification tasks. Right: Axis-wise sorting strategy adopted in segmentation tasks.}
	\label{fig:serialization_strategies}
\end{figure}

For segmentation tasks, which require local detail awareness, we propose an axis-wise sorting strategy. The center coordinates are independently sorted along the $x$, $y$, and $z$ axes:
\begin{equation}
    \begin{aligned}
    \text{idx}_x &= \text{argsort}(C_x), \\
    \text{idx}_y &= \text{argsort}(C_y), \\
    \text{idx}_z &= \text{argsort}(C_z).
    \end{aligned}
\end{equation}
and the reordered sequences are concatenated to form the final input:
\begin{equation}
    F_{\text{final}} = \text{concat}(F_{\text{idx}_x},\ F_{\text{idx}_y},\ F_{\text{idx}_z}) \in \mathbb{R}^{B \times 3G \times C}.
\end{equation}
This strategy enhances directionality and local continuity, helping the Mamba encoder better capture local geometry under its uni-directional modeling constraint. The two types of serialization strategies are visualized in Figure~\ref{fig:serialization_strategies}, where (a) shows the space-filling curve used for classification and (b) illustrates the axis-wise sorting for segmentation.

\noindent \textbf{Masking and Order Embedding.} Before feeding the sequence into the encoder, we apply either random or block-wise masking to simulate partially observed inputs. Only the visible (unmasked) tokens are embedded with positional encodings and passed into the MLA-Mamba hybrid encoder. To distinguish between different serialization strategies (e.g., Hilbert vs. axis-wise), we introduce \textit{learnable order embeddings} using the OrderScale module, which applies an affine transformation to the token features based on their ordering origin. This enhances the model’s ability to adapt to different spatial arrangements and scanning patterns.

\begin{table*}[t]\small
\setlength{\tabcolsep}{3.0pt}
\captionsetup{skip=0.1cm}
\caption{Object classification on the ScanObjectNN dataset~\cite{uy2019revisiting}. {$^\dagger$} indicates that using simple rotational augmentation~\cite{dong2022autoencoders} for training.}
\label{tab:scanobjnn}
\resizebox{1.0\textwidth}{!}{
\begin{tabular}{lcccccccc}
\toprule
Methods & Reference & Backbone & Param. (M)~$\downarrow$& FLOPs (G)~$\downarrow$ & OBJ-BG~$\uparrow$ & OBJ-ONLY~$\uparrow$ & PB-T50-RS~$\uparrow$  \\
\midrule
 \multicolumn{8}{c}{\textit{Supervised Learning Only}}\\
 \midrule
PointNet~\cite{qi2017pointnet} & CVPR 17 & - & 3.5 & 0.5  & 73.3  & 79.2  & 68.0\\
PointNet++~\cite{qi2017pointnet++}& NeurIPS 17 & - & 1.5 & 1.7 & 82.3  & 84.3 & 77.9 \\
PointCNN~\cite{li2018pointcnn}& NeurIPS 18 &- &0.6 & 0.9 & 86.1 & 85.5 & 78.5 \\
DGCNN~\cite{wang2019dynamic}& TOG 19 & - & 1.8 & 2.4 & 82.8  & 86.2  & 78.1 \\
PointNeXt~\cite{qian2022pointnext}& NeurIPS 22  & -  & 1.4 & 1.6 & -     & -  & 87.7 \\
PointMLP~\cite{ma2022rethinking}& ICLR 22  & - &  13.2 & 31.4  & -    & -     & 85.4 \\
PCM{$^\dagger$}~\cite{zhang2024point}& AAAI 25  & Mamba &  34.2 &  45  & -    & -     & 88.1 \\
PoinTramba{$^\dagger$}~\cite{wang2024pointramba}& -  & Hybrid &  19.5 & -  & 92.3    & 91.3     & 89.1 \\
 \midrule
 \multicolumn{8}{c}{\textit{Training from pre-training}}\\
 \midrule
  Point-BERT~\cite{yu2022point}& CVPR 22 &Transformer&22.1 & 4.8 & 87.43 & 88.12 & 83.07 \\
  MaskPoint~\cite{liu2022masked}& CVPR 22 & Transformer&22.1 & 4.8 & 89.30 & 88.10 & 84.30 \\
  Point-MAE~\cite{pang2022masked}& ECCV 22 &Transformer& 22.1 & 4.8 & 90.02& 88.29 & 85.18 \\
  Point-M2AE~\cite{zhang2022point}& NeurIPS 22 &Transformer& 15.3 & 3.6 & 91.22 & 88.81 & 86.43 \\
  Point-MAE+IDPT~\cite{zha2023instance}& ICCV 23 &Transformer& 1.7 & 7.2 & 91.22& 90.02 & 84.94 \\
  Point-MAE+DAPT~\cite{zhou2024dynamic}& CVPR 24 &Transformer& \textbf{1.1} & 5.0 & 90.88& 90.19 & 85.08 \\
  PointDif~\cite{zheng2024pointdif} & CVPR 24 &Transformer& - & - & 93.29 &91.91 & 87.61 \\
  Joint-MAE~\cite{guo2023joint}& IJCAI 23 &Transformer &  - & - & 90.94 & 88.86 & 86.07\\
  Point-MAE{$^\dagger$}~\cite{pang2022masked}& ECCV 22 & Transformer& 22.1 & 4.8 & 92.77 & 91.22 & 89.04 \\
  PointGPT-S{$^\dagger$}~\cite{chen2023pointgpt}& NeurIPS 23 & Transformer& 29.2 & 5.7 & 93.39 & 92.43 & 89.17 \\
  PointMamba{$^\dagger$}~\cite{liang2024pointmamba}& NeurIPS 24 & Mamba & 12.3 & \textbf{3.1} & 94.32& 92.60 & 89.31 \\
 \rowcolor{gray!30}\textbf{\ourmethod{$^\dagger$} (Ours)}& - &\textbf{Hybrid} & 12.8& 3.2 & \textbf{94.51} &\textbf{92.86} & \textbf{89.53}\\

 \bottomrule
\end{tabular}
}
\end{table*}

\subsection{Hybrid Encoder with Latent Attention}
To fully exploit the structural modeling capacity of point cloud sequences, we propose a hybrid encoder architecture that integrates Point-wise Multi-head Latent Attention (PMLA) with a Selective State Space Model (SSM), such as Mamba. Built upon the lightweight recurrence of Mamba, our design introduces a novel PMLA module to enhance local awareness, improve information flow control, and increase representational power in structurally complex point clouds.

\noindent \textbf{Hybrid Encoder Design.} Despite its impressive long-range modeling efficiency, Mamba lacks explicit local inductive bias~\cite{zhao2021point} and fails to utilize matrix multiplication units effectively~\cite{li2018pointcnn}—leading to suboptimal performance on sparse, spatially complex data such as point clouds. To address this, we introduce a customized PMLA module that replaces certain Mamba blocks in the network.

Unlike standard attention, our PMLA module removes traditional linear $QKV$ projections. Instead, it uses a 1D convolution to construct $Q$, similar to local token modeling in~\cite{han2021transformer}, preserving the order of serialized points. The $K$ and $V$ vectors are projected into low-dimensional latent spaces~\cite{jaegle2021perceiver}, gated to modulate the intensity of information flow:
\begin{equation}
K' = \text{MLP}_K(x) \odot \sigma(W_K x),\quad
V' = \text{MLP}_V(x) \odot \sigma(W_V x),
\end{equation}
The attention computation and aggregation follow:
\begin{equation}
\text{Attn}(Q, K') = \text{Softmax}\left(\frac{Q K'^T}{\sqrt{d_k}}\right),\quad
\text{Output} = \text{Attn} \cdot V'.
\end{equation}
This structure boosts semantic enhancement in local regions, particularly for spatially unordered or incomplete point sets.

\noindent \textbf{Theoretical Connection Between State Space and Latent Attention.} By comparing the update rule of selective SSM and the latent gating mechanism in PMLA, we observe conceptual alignment between both paradigms~\cite{gu2022efficiently}. The hidden state $x_t$ in Mamba reflects a compressed summary of historical features, while PMLA's latent keys and values selectively activate specific dimensions to preserve relevant information.

We further approximate the output of PMLA as a gated sample of the state accumulation in SSM:
\begin{equation}
\hat{x}_t^{\text{PMLA}} \approx \sigma(W_K x_t) \cdot \tilde{x}_t^{\text{SSM}},
\end{equation}
which reflects a controlled activation over learned latent states. In essence, PMLA serves as a learnable access regulator to the latent state space, enabling attention-like modulation within the SSM framework.

Our final encoder stacks multiple hybrid blocks combining Mamba and PMLA. Most layers remain standard Mamba blocks to capture long-range dependencies efficiently. At several key positions, we insert PMLA-enhanced blocks to boost local semantic focus and channel-wise selection. This sparsely-injected attention mechanism introduces minimal overhead but significantly improves structural perception and task generalization. The architecture of our attention block and the detailed internal design of the PMLA module are depicted in Figure~\ref{fig:mla_block}, clearly showing the residual connections and the latent attention computation.

\subsection{Conditional Diffusion for Masked Token Denoising}
As shown in Figure~\ref{fig:pipeline}, our pretraining framework adopts a conditional diffusion model that learns to denoise masked tokens, enabling structure-aware representation learning without explicit point-level supervision.

\noindent \textbf{Forward Corruption Process.} 
Given a sequence of feature tokens $\mathbf{Z}_0$, we simulate token corruption by gradually adding Gaussian noise:
\begin{equation}
    \begin{aligned}
    q(\mathbf{Z}_{1:T} | \mathbf{Z}_0) &= \prod_{t=1}^{T} q(\mathbf{Z}_t | \mathbf{Z}_{t-1}), \\
    q(\mathbf{Z}_t | \mathbf{Z}_{t-1}) &= \mathcal{N}(\mathbf{Z}_t;\sqrt{\bar{\alpha}_t}\mathbf{Z}_{t-1}, \beta_t \mathbf{I}),
    \end{aligned}
\end{equation}
where $\beta_t$ controls the noise level and $\bar{\alpha}_t = \prod_{s=1}^t (1-\beta_s)$.

\noindent \textbf{Reverse Denoising Process.} 
The goal is to recover clean tokens from noisy inputs using a denoising model conditioned on the visible tokens $\mathbf{Z}_{\text{cond}}$:
\begin{equation}
p_\theta(\mathbf{Z}_0 | \mathbf{Z}_T, \mathbf{Z}_{\text{cond}}) = \prod_{t=1}^{T} p_\theta(\mathbf{Z}_{t-1} | \mathbf{Z}_t, \mathbf{Z}_{\text{cond}}),
\end{equation}
where each step is parameterized as:
\begin{equation}
p_\theta(\mathbf{Z}_{t-1} | \mathbf{Z}_t, \mathbf{Z}_{\text{cond}}) = \mathcal{N}(\mathbf{Z}_{t-1}; \mu_\theta(\mathbf{Z}_t, \mathbf{Z}_{\text{cond}}, t), \sigma_t^2 \mathbf{I}).
\end{equation}

\noindent \textbf{Training Objective.}
We follow the standard denoising score matching objective~\cite{ho2020denoising}, minimizing the MSE between predicted and true noise:
\begin{equation}
\mathcal{L}_{\text{diff}} = \mathbb{E}_{\mathbf{Z}_0, t, \epsilon} \left[\left\| \epsilon - \epsilon_\theta(\sqrt{\bar{\alpha}_t}\mathbf{Z}_0 + \sqrt{1 - \bar{\alpha}_t}\epsilon, t, \mathbf{Z}_{\text{cond}}) \right\|^2\right].
\end{equation}

Compared with reconstruction losses like Chamfer Distance (CD)~\cite{fan2017point} or Earth Mover’s Distance (EMD), our latent-space denoising approach avoids explicit point alignment and significantly reduces computation. This makes it more scalable and compatible with sequence models like Mamba, while still capturing both local and global geometric structures through masked prediction~\cite{zhou2023pvd}.

\section{Experiments}
\subsection{Experimental Setup} 

We evaluate PointLAMA on standard point cloud benchmarks, including classification, segmentation, and ablation studies. The pretraining follows a conditional diffusion-based self-supervised framework, using consistent serialization strategies with downstream tasks to improve transferability.

Our PointLAMA encoder consists of 11 Mamba blocks and 1 Latent Attention block equipped with the proposed PMLA module. Each Mamba block operates with a feature dimension of 384, and the latent dimension in PMLA is set to 48. All experiments are conducted on a single NVIDIA RTX 4090 GPU. The backbone architecture, data processing pipeline, and training settings follow standard conventions to ensure fair comparison and reliable evaluation.

\subsection{Downstream Tasks}
We evaluate the proposed PointLAMA model on a diverse set of downstream 3D point cloud benchmarks, including real-world object classification (ScanObjectNN), clean object classification (ModelNet40), few-shot classification(ModelNet40), and part segmentation (ShapeNetPart). In each task, we compare against representative Transformer-based, Mamba-based, and hybrid architectures, with a particular focus on Mamba-only and Mamba-Transformer hybrid baselines.

\noindent \textbf{ScanObjectNN Results.} As shown in Table~\ref{tab:scanobjnn}, ScanObjectNN is a challenging dataset featuring background clutter, occlusion, and intra-class variation. PointLAMA achieves state-of-the-art performance across all three settings (OBJ-BG, OBJ-ONLY, and PB-T50-RS), notably achieving 89.53\% on PB-T50-RS, outperforming strong baselines such as PointMamba (89.31\%) and PointGPT (89.17\%).

Compared with other hybrid architectures like PoinTramba, our model introduces only a single MLA-based attention block while keeping the rest of the encoder composed of lightweight Mamba blocks. This minor complexity increase brings significant gains, highlighting the structural efficiency of our design.

\begin{table}[t]\small
\setlength{\tabcolsep}{0.5pt}
\captionsetup{skip=0.1cm}
 \caption{Classification on ModelNet40~\cite{wu20153d}. Overall accuracy (\%) is reported. The results are obtained from 1024 points without voting.}
  \label{tab:modelnet40}
  \centering
  \resizebox{0.48\textwidth}{!}{
  \begin{tabular}{lcccc}
    \toprule
    Methods & Param. (M)~$\downarrow$& FLOPs (G)~$\downarrow$ & OA (\%)~$\uparrow$  \\
    \midrule
     \multicolumn{4}{c}{\textit{Supervised Learning Only}} \\
    \midrule
    PointNet~\cite{qi2017pointnet} &3.5 &0.5 & 89.2 \\ 
    PointNet++~\cite{qi2017pointnet++} & 1.5 & 1.7  & 90.7 \\ 
    PointCNN~\cite{li2018pointcnn} &0.6 & -  & 92.2 \\ 
    DGCNN~\cite{phan2018dgcnn} &1.8 & 2.4  & 92.9 \\ 
    PointNeXt~\cite{qian2022pointnext} & 1.4 & 1.6  & 92.9 \\
    PCT~\cite{guo2021pct} & 2.9 & 2.3  & 93.2 \\
    PCM~\cite{zhang2024point}& 34.2  & -    & 90.7 \\
    PoinTramba~\cite{wang2024pointramba}& 19.5  & -    & 92.7 \\
    \midrule
     \multicolumn{4}{c}{\textit{with Self-supervised pre-training}} \\
     \midrule
      Point-BERT~\cite{yu2022point}  &22.1 & 2.3  & 92.7 \\
      MaskPoint~\cite{liu2022masked}  &22.1 & 2.3 & 92.6\\
      Point-M2AE~\cite{zhang2022point} & 12.8 & 4.6 & 93.4 \\
      Point-MAE~\cite{pang2022masked}  & 22.1 & 2.4  & 93.2 \\
      PointGPT-S~\cite{chen2023pointgpt} & 29.2 & 2.9 & 93.3 \\
      PointMamba~\cite{liang2024pointmamba}& \textbf{12.3} & \textbf{1.5}  & 93.6\\
     \rowcolor{gray!30}\textbf{\ourmethod       (Ours)} & 12.8& 1.8 & \textbf{94.5}\\
    \bottomrule
  \end{tabular}
  }
\end{table}

\begin{table}[t]\small
\setlength{\tabcolsep}{2.0pt}
\captionsetup{skip=0.1cm}
 \caption{Few-shot learning on ModelNet40~\cite{wu20153d}. Overall accuracy (\%)$\pm$the standard deviation (\%) without voting is reported.}
  \label{tab:fewshot}
  \centering
  \resizebox{0.48\textwidth}{!}{
  \begin{tabular}{lcccc}
    \toprule
    \multirow{2.3}{*}{Methods} & \multicolumn{2}{c}{5-way} & \multicolumn{2}{c}{10-way} \\
            \cmidrule{2-5}   & 10-shot & 20-shot & 10-shot & 20-shot \\
    \midrule
     \multicolumn{5}{c}{\textit{Supervised Learning Only}} \\
    \midrule
    PointNet~\cite{qi2017pointnet} & 52.0$\pm$3.8& 57.8$\pm$4.9 & 46.6$\pm$4.3 & 35.2$\pm$4.8 \\
    PointNet-CrossPoint~\cite{afham2022crosspoint} & 90.9$\pm$1.9& 93.5$\pm$4.4 & 84.6$\pm$4.7 & 90.2$\pm$2.2 \\ 
    DGCNN~\cite{wang2019dynamic} & 31.6$\pm$2.8& 40.8$\pm$4.6 & 19.9$\pm$2.1 & 16.9$\pm$1.5 \\ 
    DGCNN-CrossPoint~\cite{afham2022crosspoint} & 92.5$\pm$3.0& 94.9$\pm$2.1 & 83.6$\pm$5.3 & 87.9$\pm$4.2 \\ 
    \midrule
    \multicolumn{5}{c}{\textit{with Self-supervised pre-training}} \\
    \midrule
    Point-BERT~\cite{yu2022point} & 94.6$\pm$3.1 & 96.3$\pm$2.7 & 91.0$\pm$5.4 & 92.7$\pm$5.1 \\
    MaskPoint~\cite{liu2022masked} & 95.0$\pm$3.7 & 97.2$\pm$1.7 & 91.4$\pm$4.0 & 93.4$\pm$3.5 \\
    Point-MAE~\cite{pang2022masked} & 96.3$\pm$2.5 & 97.8$\pm$1.8 & 92.6$\pm$4.1 & 95.0$\pm$3.0 \\
    Point-M2AE~\cite{zhang2022point} & 96.8$\pm$1.8 & 98.3$\pm$1.4 & 92.3$\pm$4.5 & 95.0$\pm$3.0 \\
    PointGPT-S~\cite{chen2023pointgpt}& 96.8$\pm$2.0 & 98.6$\pm$1.1 & 92.6$\pm$4.6 & 95.2$\pm$3.4 \\
    PointMamba~\cite{liang2024pointmamba}  & 96.9$\pm$2.0 & \textbf{99.0}$\pm$1.1 & 93.0$\pm$4.4 & 95.6$\pm$3.2 \\
    \rowcolor{gray!30}\textbf{\ourmethod (Ours)} & \textbf{97.2}$\pm$1.9 & \textbf{99.0}$\pm$0.9 & \textbf{94.0}$\pm$4.5 & \textbf{95.8}$\pm$3.1 \\
    \bottomrule
  \end{tabular}
  }
\end{table}

\begin{table}[t]\small
\setlength{\tabcolsep}{0.5pt}
\captionsetup{skip=0.1cm}
 \caption{Part segmentation on the ShapeNetPart~\cite{yi2016scalable}. The mIoU for all classes (Cls.) and for all instances (Inst.) are reported.}
  \label{tab:segmentation}
  \centering
  \resizebox{0.48\textwidth}{!}{
  \begin{tabular}{lcccc}
    \toprule
    Methods   &Cls. mIoU (\%)~$\uparrow$ & Inst. mIoU (\%)~$\uparrow$ \\
    \midrule
     \multicolumn{3}{c}{\textit{Supervised Learing Only}} \\
    \midrule
    PointNet~\cite{qi2017pointnet}& 80.39 & 83.7 \\
    PointNet++~\cite{qi2017pointnet++} & 81.85 & 85.1 \\
    DGCNN~\cite{wang2019dynamic}  & 82.33 & 85.2 \\
    PoinTramba~\cite{wang2024pointramba}& -  & 85.7 \\
    PCM~\cite{zhang2024point}& 85.3  & 87.0 \\
    \midrule
    \multicolumn{3}{c}{\textit{with Self-supervised pre-training}} \\
    \midrule
    Transformer~\cite{yu2022point}   & 83.4 & 85.1 \\
    OcCo~\cite{yu2022point}  & 83.4 & 85.1 \\
    MaskPoint~\cite{liu2022masked}  & 84.6 & 86.0 \\
    Point-BERT~\cite{yu2022point}   & 84.1 & 85.6 \\
    Point-MAE~\cite{pang2022masked} & 84.2 & 86.1 \\ 
    PointGPT-S~\cite{chen2023pointgpt}  & 84.1 & 86.2 \\ 
    PointMamba~\cite{liang2024pointmamba}  & 84.4 & 86.2 \\ 
    \rowcolor{gray!30}\textbf{\ourmethod (Ours)} & \textbf{85.5} & \textbf{87.5}\\
    \bottomrule
  \end{tabular}
  }
\end{table}

\noindent \textbf{ModelNet40 Results.} As shown in Table~\ref{tab:modelnet40}, on the clean ModelNet40 benchmark, PointLAMA achieves an overall accuracy of 94.5\%, surpassing strong pre-trained models such as PointMamba (93.6\%) and PointGPT (93.3\%). This demonstrates our model's strong recognition ability and generalization even in low-noise settings.

\noindent \textbf{Few-shot Learning.} To assess generalization in low-data regimes, we conduct 5-way and 10-way few-shot classification on ModelNet40. Results in Table~\ref{tab:fewshot} show that PointLAMA consistently outperforms all existing methods, reaching 95.8\% in the 10-way 20-shot setting. This validates the strong inductive bias and transferability of our PMLA-enhanced encoder.

\noindent \textbf{Part segmentation.} As shown in Table~\ref{tab:segmentation}, we evaluate part segmentation on the ShapeNetPart dataset using the same axis-wise sorting strategy as in pre-training. PointLAMA achieves the best results among self-supervised models, reaching 85.5\% class mIoU and 87.5\% instance mIoU, surpassing PointMAE, PointMamba, and PointGPT. These results confirm the effectiveness of PMLA in modeling fine-grained local semantics.

\begin{table}[t]\small
\setlength{\tabcolsep}{2.0pt}
\captionsetup{skip=0.1cm}
 \caption{The effect of different scanning curves.}
  \label{tab:scanning_curves}
  \centering
  \resizebox{0.48\textwidth}{!}{
  \begin{tabular}{lcccc}
    \toprule
    Scanning curve & PB-T50-RS & Cls. mIoU (\%) & Inst. mIoU (\%) \\
    \midrule
     Random  & 84.45 & 82.3 & 84.3\\
     Hilbert and Trans-Hilbert & \textbf{89.53} & 84.4 &86.2 \\
     Axis-wise & 86.87 & \textbf{85.5} & \textbf{87.5} \\
    \bottomrule
  \end{tabular}
  }
\end{table}

\begin{table}[t]\small
\setlength{\tabcolsep}{2.0pt}
\captionsetup{skip=0.1cm}
\caption{The effect of including PMLA in the model.}
\label{tab:mla_inclusion}
\centering
\begin{tabular}{lcc}
    \toprule
    Setting & PB-T50-RS & OBJ-ONLY \\
    \midrule
    With MHA & 88.13 & 91.22 \\
    Only Mamba & 89.31 & 92.60 \\
    With PMLA & \textbf{89.53} & \textbf{92.86} \\
    \bottomrule
\end{tabular}
\end{table}

\begin{table}[t]\small
\setlength{\tabcolsep}{2.0pt}
\captionsetup{skip=0.1cm}
\caption{The effect of placing PMLA in the Hybrid Encoder.}
\label{tab:mla_layers}
\centering
\begin{tabular}{lccc}
    \toprule
    Layer Position & PB-T50-RS & OBJ-ONLY \\
    \midrule
    Early Layers & 88.79 & 91.88 \\
    Middle Layers & \textbf{89.53} & \textbf{92.86} \\
    Late Layers & 88.62 & 91.72 \\
    \bottomrule
\end{tabular}
\end{table}

\subsection{Ablation Study}
To evaluate the effectiveness of key components in \ourmethod, we conduct ablation studies on the PB-T50-RS and OBJ-ONLY subsets of ScanObjectNN and the ShapeNetPart segmentation benchmark. We focus on the impact of serialization strategies and the design choices of the proposed PMLA module.

\noindent \textbf{Effect of Serialization Strategy.} As shown in Table~\ref{tab:scanning_curves}, serialization plays a critical role in structural modeling. Random scanning leads to significant performance degradation (84.45 on PB-T50-RS, 82.3/84.3 mIoU), indicating the importance of structured token sequences. The Hilbert and Trans-Hilbert curves perform best for classification, while axis-wise sorting achieves optimal results on part segmentation. These findings validate the task-adaptive serialization design.

\noindent \textbf{Effect of Including PMLA.} We compare different model configurations to assess the contribution of PMLA. As shown in Table~\ref{tab:mla_inclusion}, using only Mamba blocks already achieves strong performance (89.31 / 92.60). Adding a single PMLA block further improves performance (89.53 / 92.86), demonstrating its effectiveness. Compared with full attention (MHA), PMLA offers better accuracy with fewer parameters, confirming its structural efficiency.

\noindent \textbf{Effect of PMLA Insertion Position.} Table~\ref{tab:mla_layers} investigates where to insert the PMLA block in the Hybrid Encoder (as shown in Figure~\ref{fig:pipeline}). Placing it in the middle layer yields the best performance, likely due to its ability to bridge early local and deeper global features effectively.

\noindent \textbf{Effect of Latent Dimension.} We explore different settings of the latent space dimension in PMLA (Table~\ref{tab:latent_dimension}). A dimension of 48 gives the best trade-off between capacity and generalization. Both smaller and larger settings lead to sub-optimal results, indicating that the capacity of PMLA is sensitive to this hyperparameter.

\begin{table}[H]\small
\setlength{\tabcolsep}{2.0pt}
\captionsetup{skip=0.1cm}
\caption{The effect of latent space dimension in PMLA.}
\label{tab:latent_dimension}
\centering
\begin{tabular}{ccc}
    \toprule
    Latent Dimension & PB-T50-RS & OBJ-ONLY \\
    \midrule
    32 & 88.13 & 91.26 \\
    48 & \textbf{89.53} & \textbf{92.86} \\
    64 & 88.46 & 91.53 \\
    128 & 86.76 & 89.79 \\
    \bottomrule
\end{tabular}
\end{table}

\subsection{Limitation}

Despite the strong performance of PointLAMA on various standard tasks, several limitations remain: 1) Our method relies on task-specific serialization strategies (e.g., Hilbert curves and axis-wise sorting), which may limit its generality across diverse point cloud tasks. Future work may explore more adaptive or data-driven serialization mechanisms to enhance the transferability and generalization of the model. 2) Although our method performs well on standard benchmarks, it has not yet been evaluated in more complex real-world scenarios, such as point cloud segmentation in indoor and outdoor environments, and temporally continuous 4D point cloud understanding. Extending PointLAMA to these challenging settings is an important direction for future research.

\section{Conclusion}
In this paper, we present PointLAMA, a lightweight and efficient point cloud pretraining framework that integrates the Mamba state space model with a Point-wise Multi-head Latent Attention (PMLA) module. By incorporating task-adaptive point serialization strategies and a conditional diffusion-based masked modeling scheme, PointLAMA effectively enhances structural representation while maintaining low computational overhead. Experimental results on point cloud classification, part segmentation, and few-shot learning demonstrate that PointLAMA achieves state-of-the-art performance with significantly reduced parameter count and FLOPs. More importantly, PointLAMA reveals an architectural synergy between Mamba's state space formulation and the latent-space mechanism in MLA, offering new insights into unifying state space models and attention mechanisms for scalable 3D representation learning.


\bibliography{main}

\end{document}